\documentclass[letterpaper, 10 pt, conference]{ieeeconf}
\IEEEoverridecommandlockouts    

\usepackage{multirow}
\usepackage{lipsum}
\usepackage{amsmath}
\usepackage{amssymb}
\usepackage[ruled,vlined]{algorithm2e}
\usepackage{graphicx}
\usepackage{booktabs}
\usepackage{xcolor}
\usepackage{cuted}
\usepackage{capt-of}
\usepackage{censor}
\usepackage[caption=false,font=footnotesize]{subfig}

\title{\LARGE \bf CGFM-Nav: Cognitive Graph-Field Memory for Semantic-Guided Lifelong Multimodal Embodied Navigation}

\author{Yuxiang Xiao\textsuperscript{*}, Xibei Chen\textsuperscript{*}, Xin Zhou\textsuperscript{*}, Jie Chen,
Yifeng Zhang\textsuperscript{$\dag$}, Guillaume Sartoretti
\thanks{Yuxiang Xiao, Xibei Chen, Xin Zhou, Jie Chen, Yifeng Zhang, and Guillaume Sartoretti are with the Department of Mechanical Engineering, National University of Singapore, Singapore (E-mail: \{yuxiangxiao, xibeichen, zhouxin, chen.jie, yifeng\}@u.nus.edu, guillaume.sartoretti@nus.edu.sg).}
\thanks{\textsuperscript{*} Yuxiang Xiao, Xibei Chen, and Xin Zhou contributed equally to this work.}
\thanks{\textsuperscript{$\dag$} Corresponding author: Yifeng Zhang.}
}

\begin{document}

\IEEEaftertitletext{\vspace{-1.8cm}}

\maketitle
\thispagestyle{empty}
\pagestyle{empty}


\begin{strip}
\centering
\includegraphics[width=\textwidth]{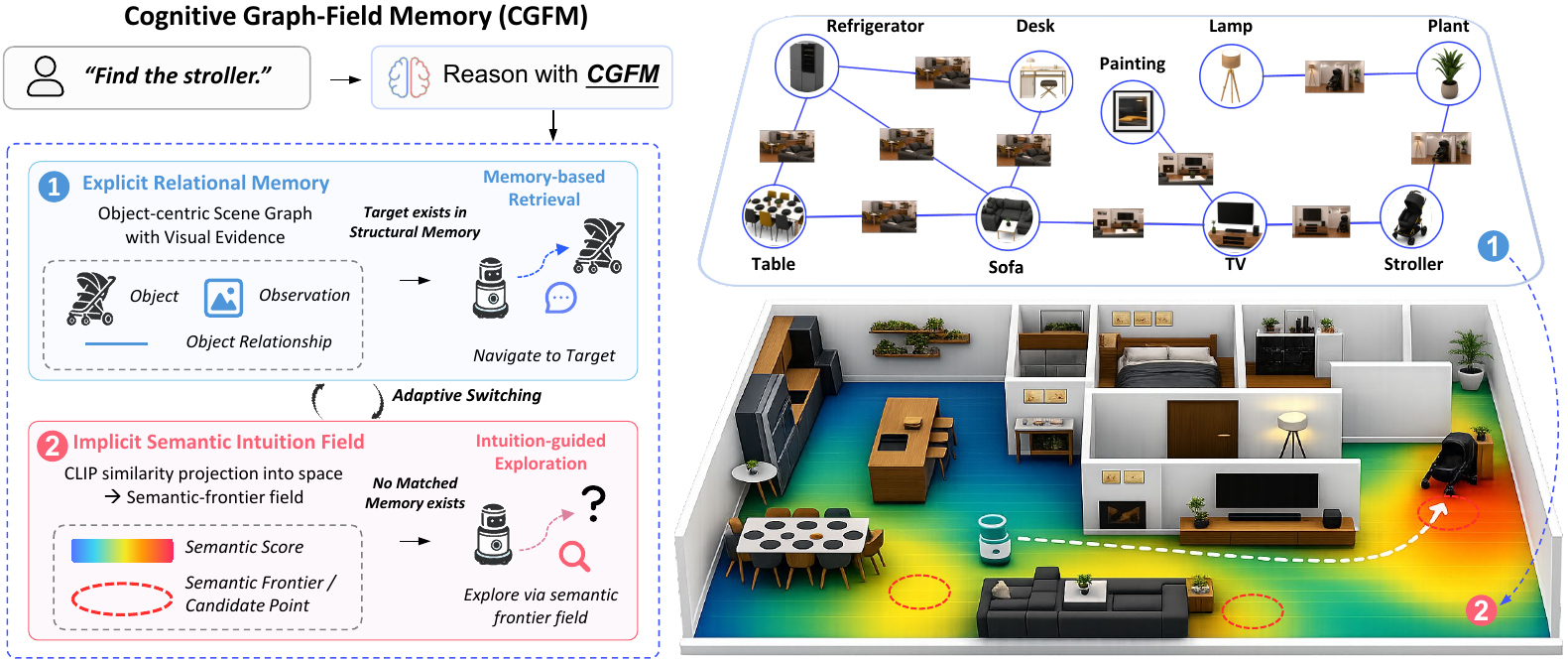}
\captionof{figure}{
Overview of the proposed Cognitive Graph-Field Memory (CGFM), inspired by the complementary mechanisms of memory and intuition underlying human navigation. (1) The multimodal scene graph serves as explicit relational memory by storing objects, spatial relations, and visual observations, enabling direct retrieval and navigation toward previously observed targets. (2) When no reliable target match is identified, graph evidence is projected into a semantic-frontier field that provides implicit spatial intuition, guiding exploration toward promising frontiers and regions.
}
\label{fig:framework}
\vspace{-0.2cm}
\end{strip}

\section{INTRODUCTION}
\label{sec:introduction}

Vision-and-Language Navigation (VLN) enables embodied agents to understand multimodal goals and navigate effectively in complex environments.
It has broad applications in assistive robotics, autonomous logistics, industrial inspection, and disaster response~\cite{gu2022vision,zhang2024vision}.
Most training-based VLN methods learn navigation behavior from task-specific supervision, which makes their performance closely tied to the training distribution but less adaptable to unseen environments and open-vocabulary goals~\cite{long2024discuss, zhou2024navgpt,zhang2024uni,zhu2025move}. 
Conversely, training-free VLN offers a promising alternative by leveraging the open-world knowledge of pretrained language and vision-language models to guide navigation without costly and complicated process of policy training~\cite{long2024instructnav,ziliotto2025tango,yin2025gc}.
In this paradigm, the agent depends heavily on how it represents the environment, specifically how visual observations, semantic cues, and spatial structure are integrated to support memory, reasoning, and decision-making.

Existing environment representations in training-free VLN can be broadly grouped into two categories. 
The first uses 2D semantic maps and frontiers, where occupancy maps and semantic heatmaps support online exploration and target search~\cite{chaplot2020object,chen2023not,yokoyama2024vlfm,yin2024sg,zhang2025apexnav}. 
These representations are efficient and easy to deploy, but they usually lack explicit object relations and long-term memory. 
The second builds scene graphs or memory graphs by organizing objects and their relations into graph structures, which enables target retrieval and relational reasoning~\cite{hughes2022hydra,gu2024conceptgraphs,werby2024hierarchical,zhou2025fsr,huang2026msgnav}.
However, such graph-based representations are often discrete and do not directly provide exploration guidance in continuous space. 
Overall, existing methods struggle to jointly achieve efficient exploration, structured memory, and geometric consistency, highlighting the need for an environment representation that supports semantic memory, continuous exploration guidance, and efficient online updates.

When navigating an unfamiliar or partially known environment, humans typically do not rely on a single form of representation~\cite{tolman1948cognitive}. 
Instead, two complementary mechanisms are often involved. One is explicit memory: for example, remembering that a chair was seen next to a table, or that a mug is likely to be found in a kitchen.
The other is a more implicit spatial intuition: even without knowing the exact location of the target, one may still have a sense that a certain direction is more likely to lead to a kitchen, or that an unexplored region is worth investigating. 
The former supports slower, structured recall and reasoning, while the latter provides a faster, less precise but efficient spatial bias.

Motivated by this cognitive perspective, we propose Cognitive Graph-Field Memory (CGFM), a unified graph-field representation that couples explicit relational memory with continuous spatial intuition, as illustrated in Fig.~\ref{fig:framework}. 
In CGFM, the scene graph serves as an explicit relational memory layer, organizing objects, spatial relations, and raw visual evidence for target retrieval and long-horizon reasoning: when a task-relevant target node exists in the graph, the agent navigates to it directly, akin to explicit recall. 
When no reliable target match can be identified, the agent instead consults a continuous semantic-frontier field derived from the same graph. By projecting visual and semantic evidence back into space, this field indicates which regions are both semantically relevant to the goal and worth exploring, resembling human spatial intuition even in the absence of explicit memory.

The graph and the field are thus not independent modules but complementary forms within the same cognitive representation: the graph's structured semantic knowledge provides the basis for field inference, while the exploration bias induced by the field in turn guides the continual expansion and updating of the graph, together forming a navigation mechanism analogous to the cooperative interaction between memory and intuition in human cognition.

Building upon CGFM, we further introduce CGFM-Nav, a foundation-model-based framework for lifelong multimodal navigation.
It incrementally collects RGB-D observations into the CGFM, extracts a compact task-relevant subgraph and combines it with decision history for VLM reasoning, and executes goal navigation or frontier/semantic exploration guided by CGFM, with verification feedback closing the loop for continual memory update.
Preliminary experiments on GOAT-Bench~\cite{khanna2024goatbench} show that, using the same Qwen3-VL-8B backbone, CGFM-Nav improves the overall success rate from 53.2\% to 63.0\% and SPL from 30.0\% to 39.6\%, while achieving consistent gains in multimodal navigation. Moreover, CGFM-Nav with Qwen3-VL-8B outperforms MSGNav~\cite{huang2026msgnav} with GPT-4o overall and on both category and language goals. 
These results suggest that coupling a persistent multimodal scene graph with a goal-conditioned semantic-frontier field enables mapping, exploration, and navigation to operate on a shared and continuously updated representation, providing an effective approach to lifelong navigation.

\section{CGFM-NAV}
\label{sec:method}

\subsection{Problem Formulation and Overview}
In this work, we consider lifelong open-vocabulary navigation with multimodal goals in an initially unknown environment, where the agent operates without task-specific policy training. 
Each episode consists of $K$ sequential subtasks, with each goal of subtask $k$ specified by an object category, a natural-language description, or a reference image, while the accumulated environment memory is retained across subtasks.
At each decision cycle $t$, the agent receives an RGB-D observation and selects a high-level navigation destination $h_t$. 
A low-level planner executes primitive actions until the target is verified or the navigation budget is exhausted.

During each navigation subtask $k$, the agent maintains
$ \mathcal{M}_t^{(k)}
=\left(
\mathcal{G}_t,
\mathcal{F}_t^{(k)},
\mathcal{D}_t^{(k)}
\right),$
where $\mathcal{G}_t$ is the persistent multimodal scene graph, $\mathcal{F}_t^{(k)}$ is the goal-conditioned semantic-frontier field, and $\mathcal{D}_t^{(k)}$ is the short-term decision memory.
Rejected candidates are not tracked separately: their verification outcomes are stored in the corresponding graph nodes and reflected in the field by suppressing their contribution. 
At the start of a new subtask, the graph and its verification history are retained, the field is reconstructed for the new goal, and the decision memory is reset.

The overall architecture and closed-loop navigation process of CGFM-Nav are illustrated in Fig.~\ref{fig:method}.

\subsection{Cognitive Graph-Field Memory}
\subsubsection{Perception and Multimodal Scene Graph Construction}

\begin{figure*}[ht]
    \centering
    \includegraphics[width=0.6\textwidth,trim=2cm 0cm 2cm 0,
    ]{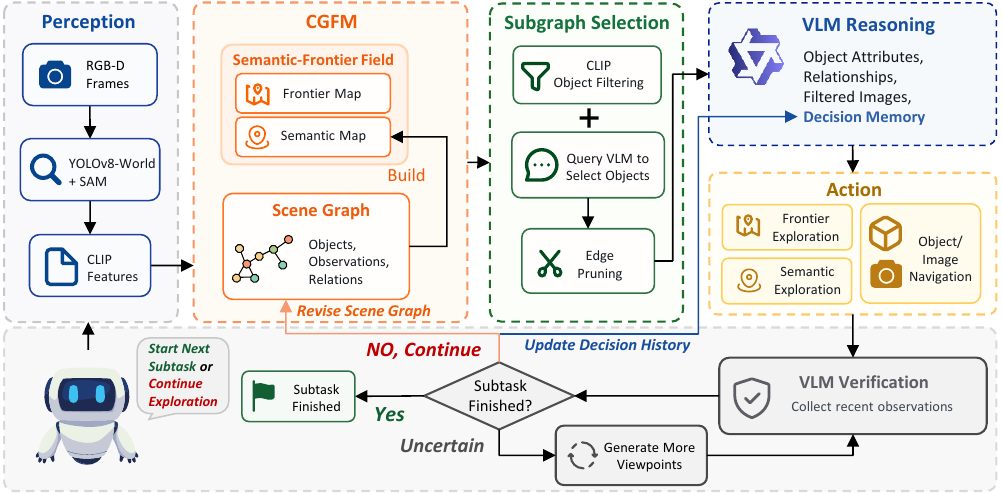}
    \caption{Overview of the proposed CGFM-Nav framework. RGB-D observations are stored in a structural multimodal scene graph, whose goal-conditioned evidence constructs a semantic-frontier field. A compact key subgraph and short-term decision memory support VLM reasoning for object/image navigation or frontier/semantic exploration. Navigation and verification outcomes update the graph, field, and decision memory, forming a closed decision loop.}
    \label{fig:method}
    \vspace{-0.5cm}
\end{figure*}

Following MSGNav~\cite{huang2026msgnav}, at each high-level cycle we process newly collected RGB-D frames using YOLOv8-World~\cite{cheng2024yoloworld} and SAM~\cite{kirillov2023segment} for open-vocabulary detection and segmentation, and extract CLIP~\cite{radford2021learning} features for target retrieval and VLM reasoning. 
These observations are incrementally associated into a multimodal scene graph $\mathcal{G}_t=(\mathcal{V}_t,\mathcal{E}_t)$, where each node $v_i=(c_i,\mathbf{p}_i,\mathbf{z}_i,\mathcal{I}_i)$ stores an object's category, 3D location, visual-semantic feature, and accumulated observation images, and each edge encodes a spatial or semantic relation between two objects.
Beyond MSGNav's base representation, we augment each node with goal-conditioned verification statistics: the verification count, associated subtask goal, final outcome, and supporting multiview feedback. 
Rather than deleting a rejected node, we suppress its contribution to the semantic field for the remainder of the subtask, preserving useful visual evidence while preventing it from repeatedly attracting the agent.

\subsubsection{Semantic-Frontier Field Construction}
The discrete scene graph is effective for retrieving known objects, but it does not directly assign exploration values to unobserved regions. 
We therefore project the semantic evidence stored in $\mathcal{G}_t$ onto a 2D navigation grid and construct a goal-conditioned dense semantic field.
For each graph node $v_i$, we compute a goal-relevance score $s_i(q)$ from its CLIP similarity to the goal, subtracting a background baseline so that only meaningfully related objects act as semantic sources, and gating it with a verification mask so that a node finally rejected for the current goal contributes nothing to the field, while an uncertain observation does not suppress it.
Each remaining object is treated as a point source, and its relevance is diffused through the observed, non-obstacle region as a weight $w_i(\mathbf{x})$ that decays exponentially with geodesic distance from the object, vanishing beyond a fixed cutoff radius. 
At each grid cell, we aggregate the $K_s$ objects with the largest local contribution into the field value
\begin{equation}
    S_t(\mathbf{x}\mid q)=
    \sum_{i\in\operatorname{TopK}_{K_s}(\mathbf{x})}
    s_i(q)w_i(\mathbf{x}).
\end{equation}

The semantic map is kept synchronized with the evolving scene graph via local updates: only cells around the previous and current projected centers of a changed object are recomputed, while the field is otherwise retained unchanged.

In parallel, the agent extracts a set of frontier candidates $\mathcal{C}_t=\{f_1,\ldots,f_M\}$ from the boundary between explored free space and unknown space, and assigns each a score $U_t(f\mid q)$ by averaging the semantic field over a small traversable neighborhood around it and discounting by its geodesic travel distance from the agent, so that closer, more semantically relevant frontiers score higher. 
If the maximum score is below a pre-defined confidence threshold, the agent falls back to the nearest frontier. 
The resulting semantic-frontier field is denoted $\mathcal{F}_t=(S_t,\mathcal{C}_t,U_t)$; unlike conventional frontier exploration, which relies mainly on geometric utility, the field $\mathcal{F}_t$ prioritizes regions that are both reachable and semantically promising for the current target.

\subsection{CGFM-Guided Navigation}

To realize zero-shot VLN through CGFM, we develop CGFM-Nav, a navigation framework built around this representation with four key modules for memory retrieval, VLM reasoning, action selection, and closed-loop refinement.

\subsubsection{Semantic-Guided Subgraph Selection}

Since goal-conditioned relevance has already been computed during field construction, CGFM reuses this evidence to replace the object-selection stage of the Key Subgraph Selection (KSS)~\cite{huang2026msgnav}, while retaining its edge-pruning and greedy image-allocation procedures.
Nodes with positive relevance and their one-hop neighbors form a semantic seed set $\mathcal{V}_{\mathrm{seed}}$; the remaining nodes are compressed into a residual graph $\widehat{\mathcal{G}}_{\mathrm{res}}$, over which a VLM performs complementary selection for candidates supported by relational or commonsense cues rather than direct CLIP similarity, described as:
\begin{equation}
    \mathcal{V}_{\mathrm{key}}
    =
    \mathcal{V}_{\mathrm{seed}}
    \cup
    \operatorname{VLMSelect}
    (\widehat{\mathcal{G}}_{\mathrm{res}},q).
\end{equation}
$\mathcal{V}_{\mathrm{key}}$ is then processed with MSGNav's edge-pruning and image-allocation procedure to form the key subgraph $\widetilde{\mathcal{G}}_t$.

\subsubsection{VLM Reasoning with Decision Memory}

The selected key subgraph provides structured environment evidence, while $\mathcal{D}_t$ maintains short-term temporal information about the current subtask.
Rather than accumulating raw VLM responses, we represent $\mathcal{D}_t$ as an ordered decision log containing the decision cycle, destination type, selected candidate identity and category, and a compact reasoning excerpt.
Before prompt construction, the log is compressed into a short list of previously visited or rejected candidates and provided to the VLM:
\begin{equation}
    d_t=
    \operatorname{VLM}
    \left(q,\widetilde{\mathcal{G}}_t,\mathcal{D}_t\right).
\end{equation}
The VLM either selects an object node $v_i$, selects an observation image $I_{i,j}\in\mathcal{I}_i$ from the key subgraph, or requests further exploration.
Accordingly, $d_t$ represents an object target, an image target, or an \textsc{Explore} request.
This decision explicitly separates navigation based on targets already represented in the scene graph from exploration guided by the semantic-frontier field.

\subsubsection{Action Selection}

For an object candidate, the stored node position is used directly as the destination; for an image candidate, YOLOv8-World and the VLM re-detect and localize the corresponding object. When the VLM requests exploration instead, the destination is chosen hierarchically: if any frontier remains ($\mathcal{C}_t\neq\varnothing$), the agent moves to the one with the highest score $U_t(f\mid q)$, falling back to the nearest frontier if all scores are weak; if no frontier remains but semantic peaks persist, it instead navigates to the centroid of the highest-response region on the semantic map; the subtask fails only when neither option is available. Exploration actions feed new observations directly into the next graph-update cycle without triggering verification.

\begin{table*}[!t]
    \centering
    \caption{Navigation performance across the first episode of each scene on the \emph{Val Unseen} split of GOAT-Bench. $\uparrow$ denotes higher is better. Best results are highlighted in bold.}
    \label{tab:results}
    \resizebox{0.9\textwidth}{!}{%
    \begin{tabular}{lcccccccc}
        \toprule
        \multirow{2}{*}{\textbf{Method}}
        & \multicolumn{4}{c}{\textbf{SR $\uparrow$}}
        & \multicolumn{4}{c}{\textbf{SPL $\uparrow$}} \\
        \cmidrule(lr){2-5} \cmidrule(lr){6-9}
        & Overall & Category & Language & Image
        & Overall & Category & Language & Image \\
        \midrule
        MSGNav (GPT-4o)       & 60.0 & 63.6 & 57.2 & \textbf{59.1} & 37.0 & 35.0 & 33.4 & \textbf{42.6} \\
        MSGNav (Qwen3VL-8B)   & 53.2   & 63.6   & 48.4   & 46.6   & 30.0   & 36.4   & 24.8   & 28.1   \\
        CGFM-Nav (Qwen3VL-8B)     & \textbf{63.0}   & \textbf{72.7}   & \textbf{61.5}   & 53.4   & \textbf{39.6}   & \textbf{44.7}   & \textbf{38.6}   & 34.8   \\
        \bottomrule
    \end{tabular}%
    }
\end{table*}

\subsubsection{Verification and Closed-Loop Update}

After reaching a navigation target, the VLM evaluates the candidate from one or more viewpoints and returns \textsc{Yes}, \textsc{No-Uncertain}, or \textsc{No-Confirmed}, completing, deferring, or rejecting the subtask accordingly; a candidate is also rejected once all viewpoints are exhausted without a positive response.
In particular, navigation outcomes and verification feedback are appended to the decision memory, and, for object candidates, also stored in the corresponding graph node, suppressing a rejected object's contribution to the semantic field for the remainder of the goal. 
This closes the loop illustrated in Fig.~\ref{fig:method}: the graph informs the field, the field and decision memory jointly determine the next action, and the resulting observation and feedback update the graph, field, and decision memory for the next cycle. 
Exploration and rejected targets return the agent to the next cycle, while a verified target advances it to the next subtask.

\section{EXPERIMENTS AND RESULTS}
\label{sec:experimental_results}

\subsection{Experiment Settings}
\label{sec:experiment_settings}

We evaluate our method on GOAT-Bench~\cite{khanna2024goatbench}, a multimodal lifelong open-vocabulary navigation benchmark comprising 360 episodes across 36 scenes, with 2669 subtasks.
Our experiments are conducted on the first episode of each scene in the \emph{Val Unseen} split, resulting in 36 evaluated episodes with 278 subtasks.
We report two key metrics: Success Rate (SR) and Success weighted by Path Length (SPL).
Given $N$ evaluated tasks, SR is defined as $\mathrm{SR}=N_{\mathrm{success}}/N$, while SPL is defined as $\mathrm{SPL}=\frac{1}{N}\sum_{i=1}^{N}S_i\frac{l_i}{\max(l_i,p_i)}$, where $S_i\in\{0,1\}$ indicates task success, $l_i$ denotes the shortest-path distance to the goal, and $p_i$ is the actual path length traveled by the agent.
We adopt the official success-distance threshold of GOAT-Bench, where a subtask is successful if the agent stops within $\mathbf{0.25\,\mathrm{m}}$ of a valid target viewpoint.

For comparison, we report the original MSGNav~\cite{huang2026msgnav} results obtained with GPT-4o (closed-source) and adopt Qwen3-VL-8B-Instruct (open-source) as the VLM backbone for MSGNav and CGFM-Nav in our experiments.

\subsection{Results}
\label{sec:results}

Table~\ref{tab:results} summarizes the multimodal navigation results on the subset from the \emph{Val Unseen} split of GOAT-Bench.
This subset covers category, language, and image goals, allowing us to evaluate CGFM-Nav across different forms of target specification.
Under the same Qwen3-VL-8B backbone, CGFM-Nav improves the overall SR from 53.2\% to 63.0\% and SPL from 30.0\% to 39.6\%, with consistent gains across all three goal modalities.
Such improvements are particularly evident for category and language goals, where navigation mainly depends on accumulating semantic evidence and efficiently exploring unseen regions.
By coupling structured scene-graph memory with a continuous semantic-frontier field, CGFM enables the agent to better reuse previous observations while guiding exploration toward semantically promising frontiers and regions.
Notably, CGFM-Nav with the lightweight, locally deployable Qwen3-VL-8B even outperforms MSGNav with the cloud-based GPT-4o in overall, category, and language performance.
This result suggests that a stronger environment representation can effectively compensate for part of the performance gap between VLM backbones, highlighting the practical potential of CGFM for efficient on-device embodied navigation.

Meanwhile, replacing GPT-4o with Qwen3-VL-8B causes a substantial performance drop on image-goal tasks for MSGNav, with SR decreasing from 59.1\% to 46.6\% and SPL from 42.6\% to 28.1\%.
CGFM improves these results to 53.4\% SR and 34.8\% SPL, but still does not surpass MSGNav with GPT-4o.
This suggests that image-goal navigation relies more heavily on fine-grained visual correspondence than semantic exploration.
While CGFM substantially improves exploration by providing more informative semantic guidance, further improvements on image-goal navigation are likely to depend on stronger visual representations or navigation-oriented VLM backbones.
\section{Conclusion and Future Work}
\label{sec:conclusion}
In this work, we introduced CGFM-Nav, a training-free framework for lifelong multimodal navigation that employs a unified graph-field representation to integrate structured scene-graph memory with semantic-guided exploration. 
Experimental results on GOAT-Bench demonstrate that CGFM-Nav consistently improves navigation performance. 
These results suggest that combining explicit semantic memory with spatial intuition provides an effective cognitive representation for embodied navigation.
In future work, we plan to evaluate CGFM on additional semantic navigation benchmarks with long-horizon and dynamic settings. We also plan to validate CGFM on a wheeled mobile robot in an office environment, enabling real-world evaluation under perception and localization uncertainties.



\newpage
\bibliographystyle{IEEEtran}
\bibliography{root} 

\end{document}